\documentclass[acmsmall,screen]{acmart}

\AtBeginDocument{%
  \providecommand\BibTeX{{%
    \normalfont B\kern-0.5em{\scshape i\kern-0.25em b}\kern-0.8em\TeX}}}

\setcopyright{acmcopyright}
\acmJournal{TIST}
\acmYear{2021} \acmVolume{1} \acmNumber{1} \acmArticle{1} \acmMonth{1} \acmPrice{15.00}\acmDOI{10.1145/3474838}

\usepackage{multirow}
\usepackage{graphicx}
\usepackage{lipsum}
\usepackage{textcomp}
\usepackage{subfigure}
\usepackage{booktabs}
\usepackage{upgreek}
\usepackage{tikz}
\usepackage{rotating}
\usepackage{pdflscape}
\usepackage{geometry}
\makeatletter
\def\ifGm@preamble#1{\@firstofone}
\appto\restoregeometry{%
  \pdfpagewidth=\paperwidth
  \pdfpageheight=\paperheight}
\apptocmd\newgeometry{%
  \pdfpagewidth=\paperwidth
  \pdfpageheight=\paperheight}{}{}
\makeatother

\begin{document}

\title{Generative Adversarial Networks for Spatio-Temporal Data: A Survey}

\author{Nan Gao}
\email{nan.gao@rmit.edu.au}
\author{Hao Xue}
\email{hao.xue@rmit.edu.au}
\author{Wei Shao}
\email{wei.shao@rmit.edu.au}
\author{Sichen Zhao}
\email{sichen.zhao@rmit.edu.au}
\author{Kyle Kai Qin}
\email{kai.qin2@rmit.edu.au}
\author{Arian Prabowo}
\email{arian.prabowo@rmit.edu.au}
\author{Mohammad Saiedur Rahaman}
\email{saiedur.rahaman@rmit.edu.au}
\author{Flora D. Salim}
\email{flora.salim@rmit.edu.au}
\affiliation{%
  \institution{RMIT University}
  \city{Melbourne}
  \state{Victoria}
  \country{Australia}
  \postcode{3000}
}

\renewcommand{\shortauthors}{Gao et al.}

\begin{abstract}
Generative Adversarial Networks (GANs) have shown remarkable success in producing realistic-looking images in the computer vision area. Recently, GAN-based techniques are shown to be promising for spatio-temporal-based applications such as trajectory prediction, events generation and time-series data imputation. While several reviews for GANs in computer vision have been presented, no one has considered addressing the practical applications and challenges relevant to spatio-temporal data. In this paper, we have conducted a comprehensive review of the recent developments of GANs for spatio-temporal data. We summarise the application of popular GAN architectures for spatio-temporal data and the common practices for evaluating the performance of spatio-temporal applications with GANs. Finally, we point out future research directions to benefit researchers in this area.
\end{abstract}



\maketitle

\section{Introduction}
Spatio-temporal (ST) properties are commonly observed in various fields, such as transportation \cite{shao2017traveling}, social science \cite{kupilik2018spatio} and criminology \cite{rumi2019crime}, among which have been rapidly transformed by the proliferation of sensor and big data. However, the vast amount of ST data requires appropriate processing techniques to build effective applications. Generally, traditional data mining methods dealing with transaction data or graph data could perform poorly when applied to ST datasets. The reasons are mainly two-fold  \cite{wang2020deep}: (1) ST data are often in continuous space while traditional data (e.g., transaction data, graph data) are usually discrete; (2) ST data usually have spatial and temporal attributes where the data correlations are more complex to be captured by traditional techniques. Moreover, ST data tend to be highly self-correlated, and data samples are usually not generated independently as in traditional data. 

With the prevalence of deep learning, many neural networks (e.g., \textit{Convolutional Neural Network} (CNN) \cite{krizhevsky2012imagenet}, \textit{Recurrent Neural Network} (RNN) \cite{mikolov2010recurrent}, \textit{Autoencoder} (AE) \cite{hinton2006reducing}, \textit{Graph Convolutional Network} (GCN) \cite{kipf2016gcn}) have been proposed and achieved remarkable success for modelling ST data, due to its demonstrated potential for hierarchical feature engineering ability. However, the traditional deep learning based ST modelling methods have some limitations. For instance, existing methods use deterministic models (e.g., RNN) and cannot capture the stochastic behaviour of ST data. Additionally, traditional deep learning approaches lack effective mechanisms to support the reasoning of the abstract data, which makes it hard to identify the factors leading to model improvements \cite{saxena2019d}. To address the above challenges, we have explored  one of the most interesting breakthroughs in the deep learning field: \textit{Generative Adversarial Networks} (GANs) \cite{goodfellow2014generative}, which can learn rich distributions over ST data implicitly and work with multi-model outputs \cite{saxena2019d}. 

GAN is a generative model which learns to produce realistic data adversarially. It consists of two components  \cite{goodfellow2014generative}: the generator $G$ and discriminator $D$. $G$ captures the data distribution and produces realistic data from the latent variable $z$, and $D$ estimates the probability of the data coming from the real data space. GAN adopts the concept of the zero-sum non-cooperative game where $G$ and $D$ are trained to play against  each other until reaching a Nash equilibrium. Recently, GANs  have gained considerable attention in various fields \cite{saxena2020generative, gui2020review}, involving images (e.g., image translation \cite{isola2017image}, super-resolution \cite{ledig2017photo}, joint image generation \cite{liu2016coupled}, object detection \cite{ehsani2018segan}, change facial attributes \cite{donahue2017semantically}), videos (e.g., video generation \cite{chai2020crowdgan,chen2020scripted,chu2020learning,wang2020imaginator}, text to video \cite{ijcai2019-276}), and natural language processing (e.g., text generation \cite{lin2017adversarial}, text to image \cite{zhang2017stackgan}). 


However, image or video generation approaches are not applicable for modelling traditional ST data (e.g., time series, trajectories, ST events, ST graphs) in real-world applications such as traffic flow, regional rainfall, and pedestrian trajectory. On the one hand, image generation usually takes the appearance between the input and output images into account, and fails to adequately handle spatial variations. On the other hand, video generation considers spatial dynamics between images, however, temporal changes are not adequately considered when the prediction of the next image is highly dependent on the previous image \cite{saxena2019d}. Though the video can be regarded as a special type of ST data due to its  dynamic locations in spatial and temporal dimensions, the discussion of using GANs for video generation usually falls into the field of computer vision, where several papers have thoroughly reviewed the recent progress of video generation with GANs \cite{wang2021generative,liu2021gene}. Hence, new approaches need to be explored to successfully modelling ST data with GAN techniques.

Recently, GANs have been applied to ST data modelling, where the applications usually include the generation of de-identified ST events \cite{saxena2019d, jin2019crime}, time series imputation \cite{luo2018multivariate,luo20192}, trajectory prediction \cite{gupta2018social, kosaraju2019social}, graph representation \cite{wang2018graphgan, bojchevski2018netgan}, etc. Despite the success of GANs in the computer vision area (e.g., image and video generation), applying GANs to ST data prediction is challenging \cite{saxena2019d}. For instance, leveraging additional information such as \textit{Places of Interest (PoI)} and weather information is still untouched in previous research. Besides, different from the images where researchers could rely on visual inspections of the generated instances, evaluation of GANs on ST data remains an unsolved problem. It is neither practical nor appropriate to adopt the traditional evaluation metrics for GAN on ST data \cite{saxena2019d,esteban2017real}.

A few studies reviewed recent literature on ST data modelling problems or GAN based applications in different fields. For ST data modelling, Atluri et al. \cite{atluri2018spatio} reviewed the popular problems and methods for modelling ST data. A taxonomy of the different types of ST data instances has been provided to identify the relevant problems for ST data in real-world applications. Then, Wang et al. \cite{wang2020deep} reviewed the recent progress in applying deep learning to ST data mining tasks and proposed a pipeline of the utilisation of deep learning models for ST data modelling problems. For GAN based applications, Hong et al. \cite{hong2019generative} explained the GANs from various perspectives and enumerated popular GAN variants applied to multiple tasks. Recent progress of GANs was discussed in \cite{pan2019recent} and Wang et al. \cite{wang2021generative} proposed a taxonomy of GANs for the computer vision area. Particularly, Yi et al. \cite{yi2019generative} reviewed the recent advances of GANs in medical imaging. 

Nevertheless, all the above works reviewed either ST data modelling problems or the recent progress of GANs in the computer vision area \cite{saxena2020generative, gui2020review}. Though many researchers \cite{saxena2019d,esteban2017real,gupta2018social,luo20192,luo2018multivariate} have modelled ST data with GANs, there is no related survey in this area to address the potential of using GANs for ST data applications. The lack of a comprehensive review makes it more difficult for researchers to identify the problems and choose an appropriate method (e.g., architecture, loss function, evaluation metric) when applying GAN techniques for ST applications. For the first time, this paper presents a comprehensive overview of GANs for ST data, describes promising applications of GANs, and identifies some remaining challenges needed to be solved for enabling successful applications in different ST related tasks.

To present a comprehensive overview of all the relevant research on GANs for ST data, we use \textit{Google Scholar} \footnote{https://scholar.google.com/} to conduct automated keyword-based search \cite{rowe2014literature}. According to \cite{al2012intelligent}, Google Scholar provides coverage and accessibility, and digital libraries such as \textit{IEEE Explore} \footnote{https://ieeexplore.ieee.org/}, \textit{Science Direct} \footnote{https://www.sciencedirect.com/}, \textit{ACM Digital Library} \footnote{https://dl.acm.org/}. 
The search period is limited from 2014 to 2021 (inclusive) as the GAN has first appeared in 2014 \cite{goodfellow2014generative}. However, papers that introduce novel concepts or approaches for ST data mining can be predated 2014. To ensure that our survey covers all relevant primary literature, we have included such seminal papers regardless of their publication date.

The remainder of the paper is organised as follows. In Section \ref{sec:pre}, we discuss the properties, characteristics and common research problems of ST data. We also present the popular deep learning methods with non-GAN frameworks for ST data, including the \textit{Convolutional Neural Networks}, \textit{Recurrent Neural Networks}, \textit{Long Short-term Memory} and \textit{Gated Recurrent Units}. Section \ref{sec:gan} reviews the definition of GAN and its popular variants with different architecture and loss functions. Section \ref{sec:st} lists the recent research progress for GANs in different categories of ST  applications. Section \ref{sec:dis} summarises the challenges of processing ST data with GANs, including the adapted architectures, loss functions and evaluation metrics. Finally, we conclude the paper and discuss future research directions.

\section{Preliminary}
\label{sec:pre}
\subsection{Spatio-temporal Data}
The existence of time and space introduces a wide variety of ST data types, leading to different ways of formulating ST data mining problems and techniques. In this part, we will first introduce the general properties of ST data, then describe the common types of ST data in different applications using generative adversarial nets techniques.
\subsubsection{Properties}
There are several general properties for ST data (i.e., spatial reference, time reference, auto-correlation, and heterogeneity \cite{atluri2018spatio}) described as below.

\textbf{Spatial Reference}. The spatial reference describes whether the objects are associated with the fixed location or dynamic locations \cite{kisilevich2009spatio}. Traditionally, when the data is collected from stationary sensors (e.g., weather stations), we consider the spatial dimension of the data is fixed. Recently, with the boost of mobile computing and location-based services, the dynamic locations of moving objects have been recorded where the collected data comes from sensors attached to different objects, e.g., GPS trajectories from road vehicles \cite{prabowo2019coltrane}. 

\textbf{Temporal Reference}. The temporal reference describes to what extent the objects evolve \cite{kisilevich2009spatio}. The simplest context includes objects that do not evolve where only the static snapshots of objects are available. In a slightly more complicated situation, objects can change status but only the most recent update snapshot remains where the full history of status is unknown. The extreme context consists of moving objects where the full history of moving is kept, therefore generating time series where all the status have been traversed. 

\textbf{Auto-correlation}. The observations of ST data are not independent and usually have spatial and temporal correlations between near measurements. For example, in the transportation area, sensors in each parking lot with the unique spatial location can record the temporal information when a vehicle arrives or leaves \cite{shao2017traveling}. This auto-correlation of ST data results in the smoothness of temporal measurements (e.g., temperature changes over time) and consistency between the spatial measurements (e.g., temperature values are similar in adjacent locations). Thereby, the traditional GAN techniques for the computer vision field (e.g., image generation \cite{goodfellow2014generative}) without considering the temporal correlation may not well suited for the ST data. 

\textbf{Heterogeneity}. ST dataset can show heterogeneity in spatial or temporal information on different levels. For instance, traffic flow in a city can show similar patterns between different weeks. During a week, the traffic data on Monday may be different from data on Friday. There can also be inter-week changes due to public events or extreme weather, affecting the traffic patterns in a city. To deal with the heterogeneity of spatial and temporal information, it is necessary to learn different models for different spatio-temporal regions \cite{bhatia2020exgan}. 

\subsubsection{Data Types}
There are various spatio-temporal data types in real-world applications, differing in the representation of space and time context \cite{atluri2018spatio}. Hence, it is crucial to establish the available ST data types in applications to effectively use GAN methods. Here, we describe the four common types of ST data which have been studied with GANs recently: (1) time series \cite{mogren2016c,esteban2017real,hartmann2018eeg,chen2018building,che2017boosting,luo2018multivariate,luo20192,li2019mad,koochali2019probabilistic,zhou2018stock}; (2) ST events \cite{saxena2019d,shao2017traveling}; (3) ST graphs \cite{lei2019gcn, yang2019advanced,wang2018graphgan}; (4) trajectory data \cite{gupta2018social}. Among the above four types of ST data, ST events and trajectories capture the observations of discrete objects and events. At the same time, the time series and ST graphs record the information of continuous or discrete ST fields. Though there are other types of ST data available in real-world scenarios, in some cases they can be converted into another, or they can be processed with similar GAN approaches to the above four types (e.g., sequential data vs time series).  Next, we briefly discuss the properties of those data types and potential difficulties when facing with GANs.

\textbf{Time Series}.
A time series can be represented as a sequence of data points $X=\{X_1,X_2,...,X_n\}$ listed in an order of time (i.e., sequence of discrete-time data \cite{tretter1976introduction}). Examples of time series include the values of indoor temperature during a day \cite{gao2020transfer,rahaman2020ambient,gao2021understanding}, the changes of accelerometer readings in the IoT devices \cite{gao2019predicting,gao2020n}, fluctuations of the stock price in a month \cite{zhou2018stock}, etc. Time series analysis consists of techniques to analyse time series for extracting useful statistic information and data characteristics. The common questions used for dealing with time series include but not limited to:
\textit{Can we predict future values for time series based on the historical values \cite{muller1997predicting, weigend2018time,koochali2019probabilistic}?}
\textit{Can we cluster groups of time series with similar temporal and spatial patterns  \cite{aghabozorgi2015time,liao2005clustering}?}
\textit{Can we impute the missing values automatically in multi-variate time series \cite{moritz2017imputets,luo20192}?}
\textit{Can we split time series into different segments with its characteristic properties \cite{jamali2015detecting, deldari2020espresso}?} 

\textbf{Spatio-temporal Events}.
\begin{figure}
    \centering
    \subfigure[Spatio-temporal events]{ \includegraphics[width=0.35\textwidth]{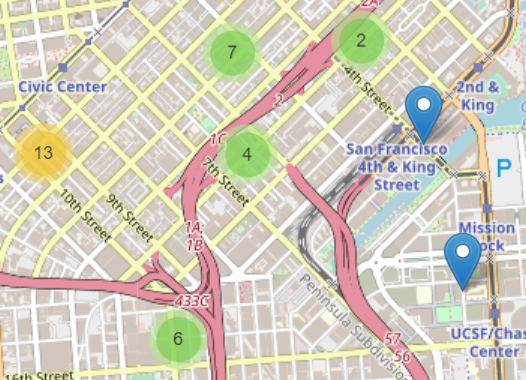}\label{fig:STevents}}\hspace{0.6cm}
    \subfigure[Two trajectories]{\includegraphics[width=0.43\textwidth]{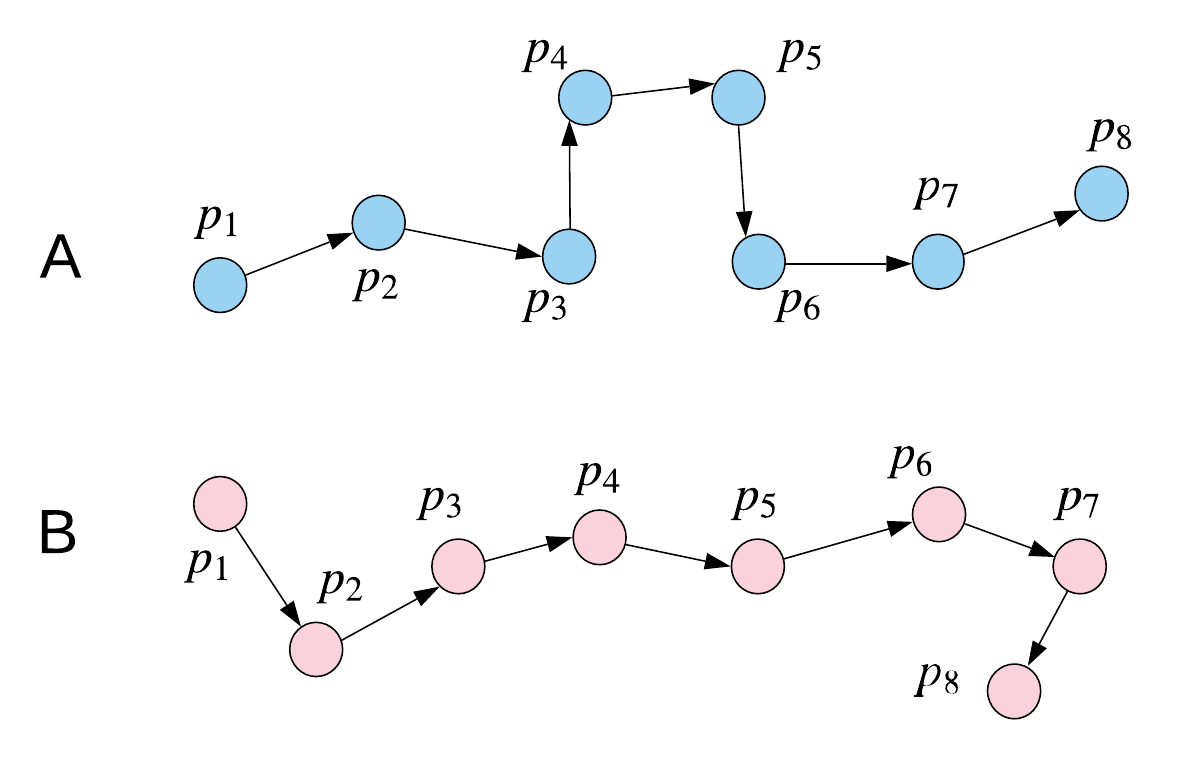}
    \label{fig:Traj}}
    \caption{Examples of spatio-temporal events and trajectories}
\end{figure}
An spatio-temporal event represents a tuple containing temporal, spatial information as well as an additional observed value \cite{li2013spatio}. Generally, it is denoted as $x_i = \{m_i,t_i,l_i\}$, where $t_i$ and $l_i$ indicates the time and location of the event, $m_i$ means the value to describe the event. Typically, the locations are recorded in three dimensions (i.e., latitude, longitude, and altitude or depth),  although sometimes only 1 or 2 spatial coordinates are available. Spatio-temporal events (see Figure ~\ref{fig:STevents}) are frequently used in real-world applications such as the taxi demand \cite{rahaman2017predicting}, traffic flow \cite{saxena2019d}, urban crimes \cite{rumi2018crime}, forest fires \cite{de2009omg}, etc. In some cases, spatio-temporal events may even have duration like parking or heliophysics \cite{pillai2013filter}. Usually, an ordered set of spatio-temporal events can also be considered as an trajectory where the spatial locations visited by moving objects. Some common questions that used for analysing spatio-temporal events includes: \textit{Can we predict the future spatio-temporal events based on the previous observations \cite{saxena2019d}?} \textit{How are spatio-temporal events clustered based on time and space \cite{shao2016clustering}?} \textit{Can we identify the anomalous spatio-temporal events that do not follow the common patters of other events \cite{barz2018detecting}?}


\textbf{Trajectory data}.
A trajectory represents the recordings of locations of a moving object at certain times and it is usually defined as a function mapped from the temporal domain to the spatial domain \cite{Frentzos2009,bian2019trajectory}. Trajectories of moving points can be denoted as a sequence of tuples
$P =\{(x_1,y_1,t_1),(x_2,y_2,t_2),...,(x_n,y_n,t_n)\}$, where $(x_i,y_i,t_i)$ indicates the location $(x_i,y_i)$ at time $t_i$. Several research have been conducted in the field of trajectory data mining and there are four major categories \cite{zheng2015trajectory}: mobility of people \cite{ren2017d,spaten}, mobility of transportation \cite{saxena2019d}, mobility of natural phenomena and mobility of animals \cite{li2010mining}. Figure ~\ref{fig:Traj} shows an example of two trajectories of object $A$ and object $B$. The common questions for processing trajectory data include:  
\textit{Can we predict the future trajectory based on the historical trajectory traces \cite{gupta2018social,sadri2017full,sadri2018will}?} 
\textit{Can we divide a collection of trajectories into small representative groups \cite{shao2019onlineairtrajclus}?}
\textit{Can we detect the abnormal behaviours from trajectories \cite{liu2013fraud}? } 

\textbf{Spatio-temporal Graph}.
\begin{figure}
    \centering
    \includegraphics[width=0.9\textwidth]{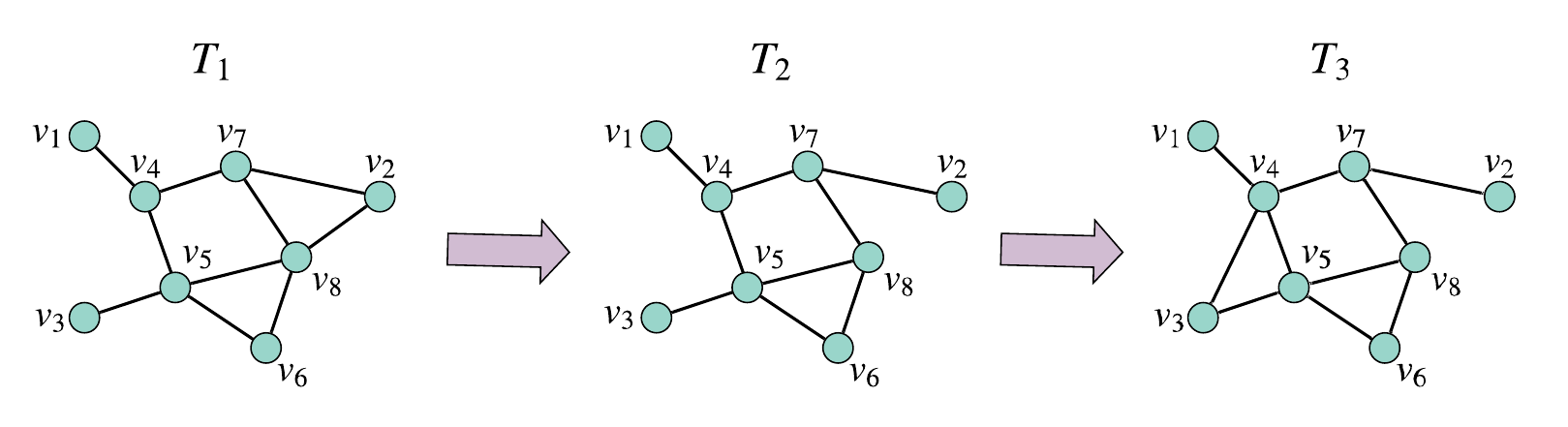}
    \caption{Example of Spatio-temporal Graph Data}
    \label{fig:graph}
\end{figure}
Spatio-temporal graph structure provides the representation of the relations between different nodes in different time.  A sequence of spatio-temporal graphs \cite{yang2019advanced} can be represented as $\mathcal{G}=(\mathcal{G}_1, \mathcal{G}_2,...,\mathcal{G}_n)$ where $\mathcal{G}_i = \{V_i, E_i, W_i\}$ indicates the graph snapshot at time $T_i$ ($i\in \{1,2,...,n\}$). Spatio-temporal graphs have been applied in various domains such as commerce (e.g., trades between countries \cite{ma2017comparative}), transportation (e.g., route planning algorithms \cite{george2007spatio}, traffic forecasting \cite{yu2018spatio}) and social science (e.g., studying geo-spatial relations of different social phenomena \cite{Gunturi2018}). Figure~\ref{fig:graph} is an example of spatio-temporal graphs in $T_1, T_2, T_3$. Some common questions for processing spatio-temporal graph includes:
\textit{Can we forecast the status of graph based on the historical graph representations \cite{wang2018graphgan,yu2018spatio}? }
\textit{Can we predict the links based on the previous graph networks \cite{lei2019gcn}? }

\subsection{Spatio-Temporal Deep Learning with Non-GAN Networks}
Here we introduce the traditional deep learning approaches for ST data with non-GAN networks (i.e., \textit{Convolutional Neural Network}, \textit{Recurrent Neural Network}, \textit{Autoencoder}, \textit{Graph Convolutional Network}), which are usually integrated into GAN architectures in ST data modelling.
\subsubsection{CNN}
\textit{Convolutional Neural Network} (CNN) \cite{krizhevsky2012imagenet} is a type of deep, feed-forward neural network commonly used to analyse visual imagery. A typical CNN model is composed of an input layer, an output layer and some hidden layers. 
Compared to the traditional multilayer perceptron (MLP), CNNs can develop internal representations of two-dimensional images, allowing CNNs to be used more generally on other types of data with spatial correlations. Though CNNs are not specifically developed for non-image data, it has been widely used in ST data mining problem for trajectory and ST raster data \cite{prabowo2019coltrane}.

\subsubsection{RNN, LSTM and GRU}

\begin{figure}%
	\centering
	\subfigure[RNN]{%
	\label{fig: rnn}%
	{\includegraphics[height=1.1in]{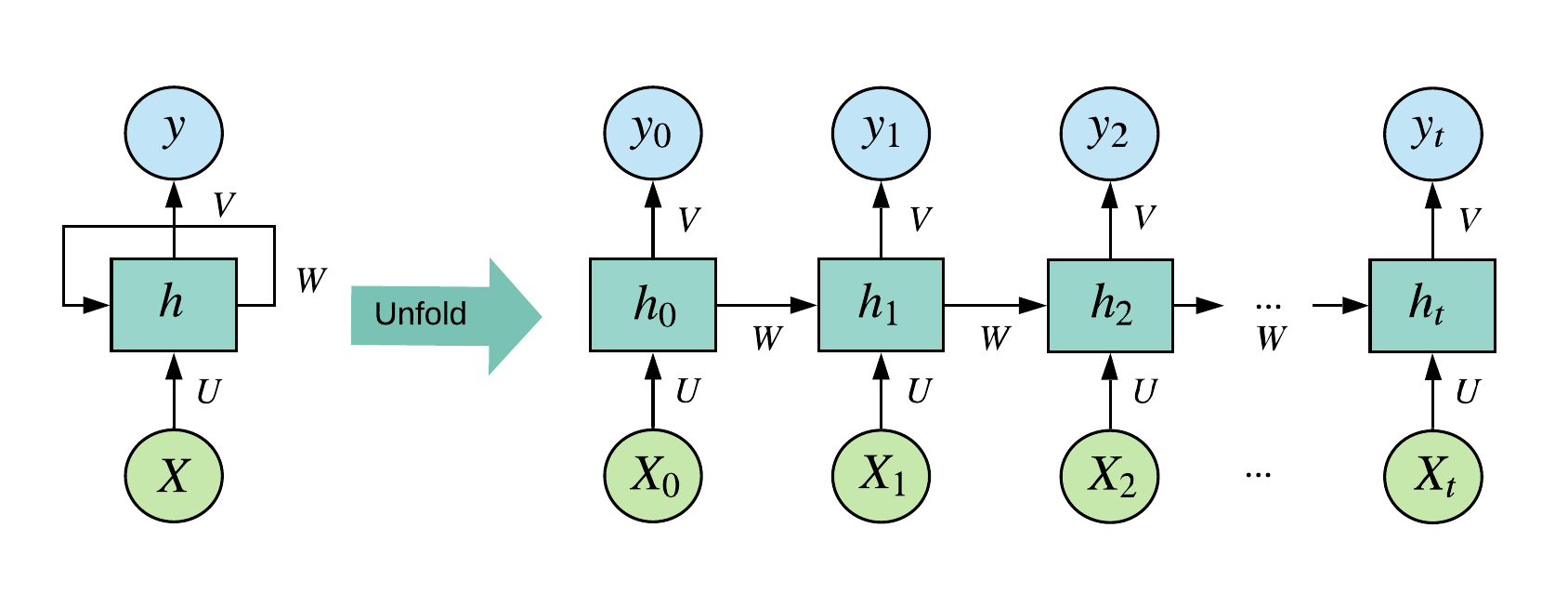}}%
	}
	\hspace{0.15in}
	\subfigure[Autoencoder]{%
	\label{fig:autoencoder}%
	\includegraphics[height=1.1in]{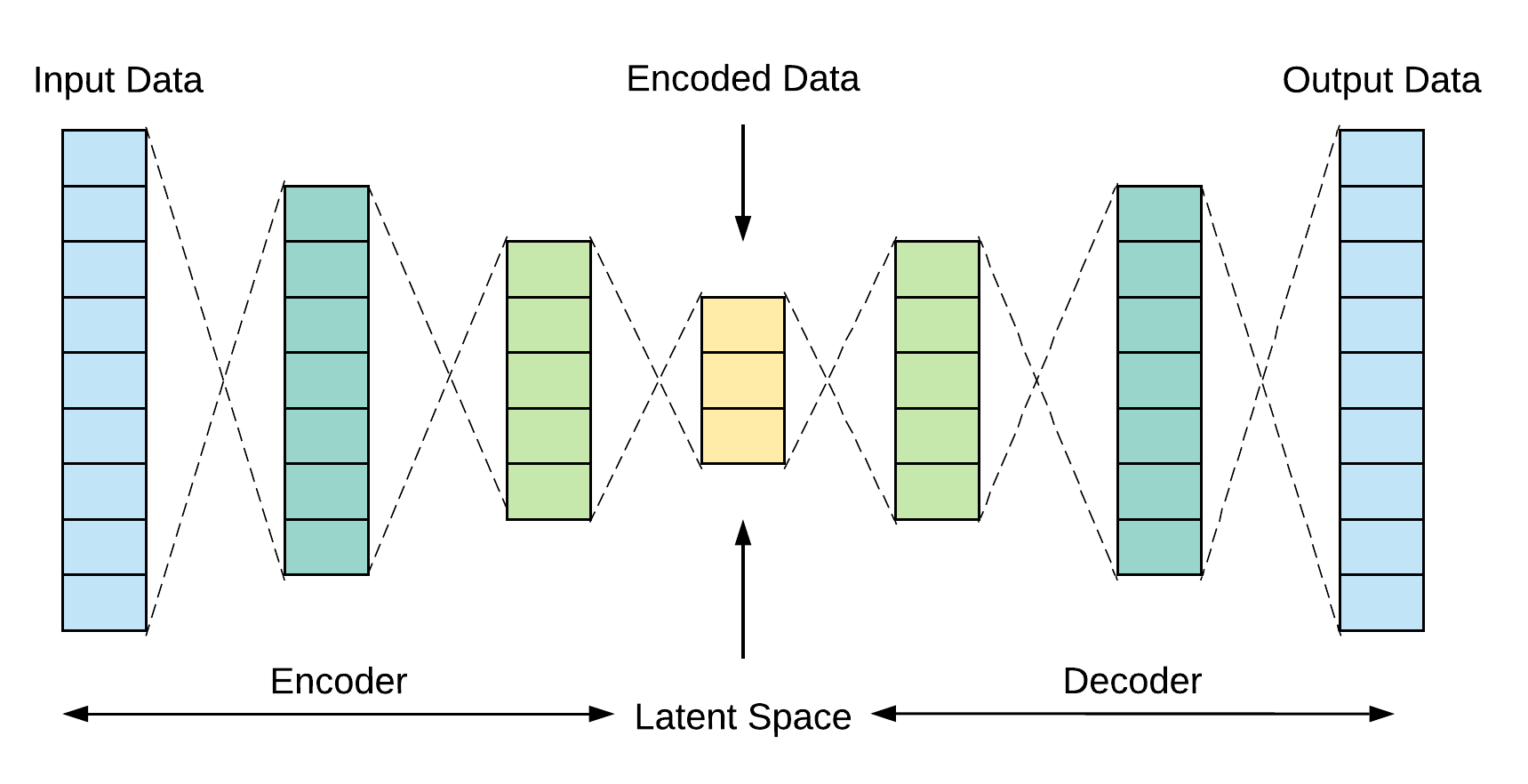}%
    }
    \caption{Structure of RNN and Autoencoder }
\end{figure}

\textit{Recurrent Neural Network (RNN)} \cite{mikolov2010recurrent} is a type of neural networks where the previous outputs are fed as the input to the current step.
The advantage of RNN is the hidden state (internal memory) that captures information calculated so far in a sequence. Figure~\ref{fig: rnn} shows the basic architecture of an RNN, where $X$ is the input data, $y$ is the output data, $h$ is the hidden state and $U, V, W$ indicates the parameters of the RNN. The current state $h_t$ is calculated by the current input $X_t$ and previous state $h_{t-1}$. 

Though the RNNs work effectively in many application domains, they may suffer from a problem called vanishing gradients \cite{li2018independently}. To cope with this problem, two variants of RNN have been developed: Long Short-Term Memory (LSTM) \cite{hochreiter1997long} and Gated Recurrent Units (GRU) networks \cite{cho2014properties}. LSTM is capable of learning long-term dependencies with a special memory unit. An LSTM cell has three gates (forget gate, input gate, and output gate) to regulate the information flow. 
Compared with standard LSTM models, GRU has fewer parameters which combines the input gate and the forget gate into an 'update gate' and merges the cell state and hidden state. RNN, LSTM and GRU are widely used to learn the temporal correlations of time series and ST data.

\subsubsection{Autoencoder (AE)}
AE \cite{hinton2006reducing} is a neural network that is trained to copy its input to its output by learning data codings in an unsupervised manner \cite{goodfellow2016deep}. The network is composed of two parts: encoder and decoder, as shown in Figure~\ref{fig:autoencoder}. The encoder function compresses the input into a latent-space representation and the decoder reconstructs the input through the representation. 
As a commonly used unsupervised representation learning method, AE is popular for classification and prediction tasks in trajectories \cite{nguyen2012extracting,zhou2018trajectory}, time series \cite{hossain2015forecasting} and other ST data \cite{duan2014deep}.


\subsubsection{Graph Convolutional Network (GCN)} 
With the ability to extract representations from both local graph structure and node features, GCN \cite{kipf2016semi} has become popular in solving learning tasks on spatio-temporal graph dataset. For instance, Yu et al. introduced Spatio-Temporal Graph Convolutional Networks (STGCN) \cite{yu2018spatio} to solve the prediction problem in traffic networks. Other deep learning models have their issues dealing with ST forecasting tasks, such as RNN-based networks often have heavy computation in training and normal convolutional operations are limited on grid structures. STGCN differently converts traffic data into the graph-structured format and use spatio-temporal convolutional blocks to capture spatial and temporal dependencies. Furthermore, the cost of computation could be reduced by Chebyshev Polynomials Approximation or First Order Approximation. Recently, attention mechanisms have been employed with GCN models to learn the impact of the spatio-temporal factors in training, such as Graph Multi-Attention Network (GMAN) \cite{zheng2020gman} and Attention-based Spatial-Temporal Graph Convolutional Network (ASTGCN) \cite{guo2019attention}.

\section{Generative Adversarial Networks}
\label{sec:gan}

\subsection{Basic Idea of GANs}

\begin{figure}
    \centering
    \includegraphics[width=0.95\textwidth]{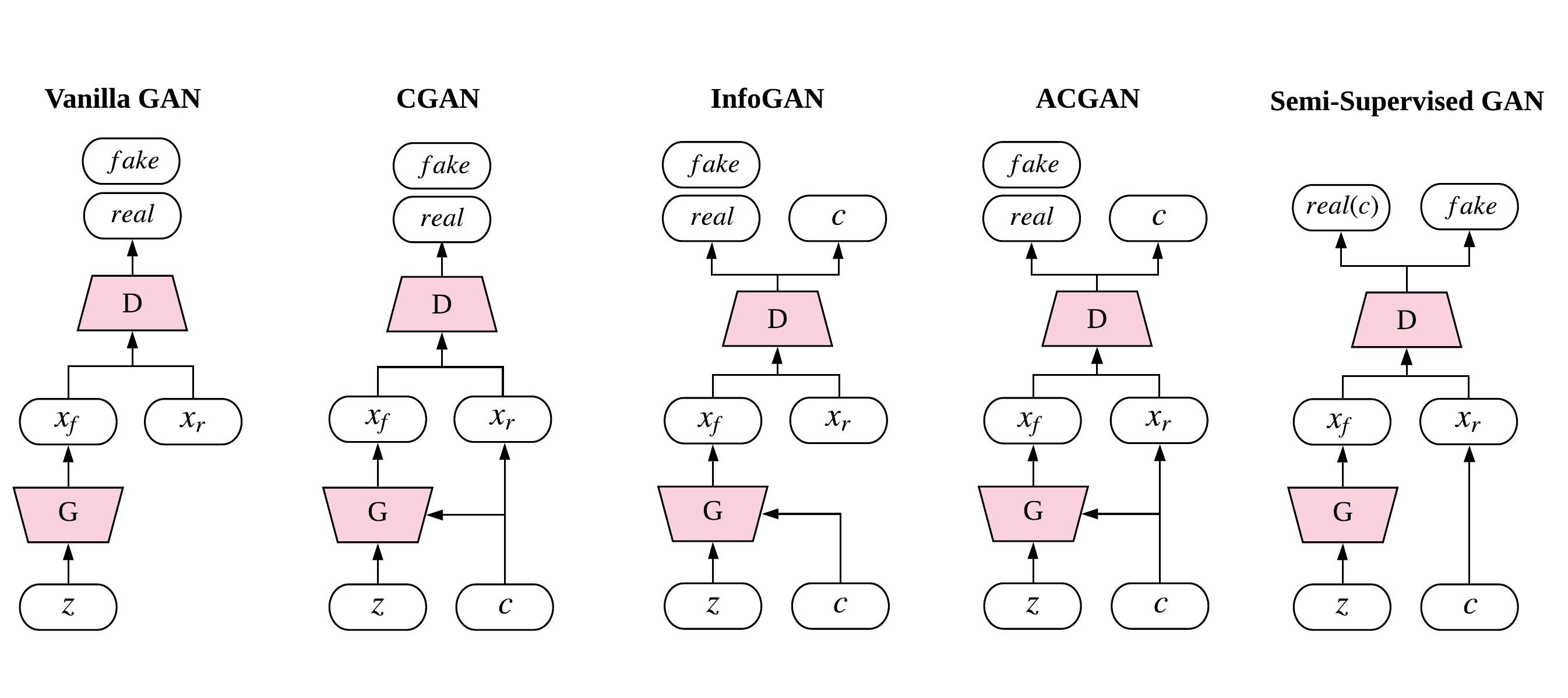}
    \caption{A view of variants of GAN. $G$ represents the generator network, $D$ is the discriminator network, $z$ represents the noise, $c$ means the class labels, $x_f$ are the generated fake images and $x_r$ are the real images}
    \label{fig:variants}
\end{figure}

The original concept of GANs is to create two neural networks and let them compete against each other. As shown in Figure~ \ref{fig:variants}, the basic architecture of GANs comprises two components: a generator and a discriminator. On the one hand, the generator's task is to synthesis fake images which can fool the discriminator. On the other hand, the discriminator, as to its name, learns to distinguish if its input is a fake image or not \cite{goodfellow2014generative}. 

Let's left the images generation task aside. The underlying idea of generative adversarial nets is more general, which is to create one fake distribution $p_g$ and make it as close as possible to a data distribution $p_r$. The reason we use such an approach is that $p_r$ could be hard to get directly and by doing in this manner, we get a good approximation of it and then we can sample from this approximate distribution instead \cite{arjovsky1701towards}. The advantages of this approach are that since the generator is learning to approximate the real distribution directly, there is no need to introduce the Markov Chain and no inference is required due to the isolation between the generator and the real data distribution. Besides, its simple structure makes it easier to incorporate with other techniques \cite{cgan}.

The Generator $G(z;\theta_g)$, a neural network that parameterized by theta takes a sample $z \sim p_z$ as input and mapping that to a sample $x\sim p_g$. And its rival, the Discriminator $D(x;\theta_d)$ outputs a single binary value to indicate its prediction of the input's origin. During the training session, both parts are trained simultaneously and based on their opponent's result, which forms a minimax game with the overall objective function \cite{goodfellow2014generative}:

\begin{equation*}
\underset{G}{min}\underset{D}{\ max} \ V( D,G) =\mathbb{E}_{x\sim p_{data}( x)}[\log D( x)] +\mathbb{E}_{z\sim p_{z}( z)}[\log( 1-D( G( z)))]
\end{equation*}

Despite all the advantages above, the original generative adversarial network is still inadequate in some places. The practical results show that the training is particularly delicate and the generators may suffer from vanishing gradient for optimizing the generator \cite{arjovsky1701towards}.  To address all those problems that might occur, many variants of the vanilla GAN are proposed \cite{acgan,cgan,infogan,stackgan}. 

\subsection{Loss Function}
In traditional generative modelling approaches, the performance of a model is indicated by the reverse Kullback-Leibler (KL) divergence between our desire distribution $p_r$ and our generator's distribution $p_g$ \cite{arjovsky1701towards}. 

\begin{equation*}
D_{KL}( P_{g} \| P_{r}) =\int _{\chi } P_{g}( x) \ \log\frac{P_{g}( x)}{P_{r}( x)}\mathrm{d} x
\end{equation*}

Minimising this term means making those two distributions closer, and it would get to zero once $p_r=p_g$. However, the generator might still generate fake-looking data due to the imbalanced nature \cite{arjovsky1701towards} of this function. It could heavily penalise the generator for the part that is in the real distribution but not covered by the generator while paying less attention to the extra part covered by the generator. To avoid this weakness,  another option that is discussed in the original GANs paper is called Jensen-Shannon (JS) divergence \cite{goodfellow2014generative}. 

\begin{equation*}
D_{JS}( P_{r} \| P_{g}) =\frac{1}{2} D_{KL}\left( P_{r} \| \tfrac{1}{2}( P_{r} +P_{g})\right) +\frac{1}{2} D_{KL}\left( P_{g} \| \tfrac{1}{2}( P_{r} +P_{g})\right)
\end{equation*}

Although it shows some promising results, JS divergence is not the ultimate choice since it still suffers from issues like gradient vanishing. Some latest studies show that those can be resolved by using other types of loss function \cite{wgan, fgan, lsgan}.

\subsection{Architecture of GAN Variants}
Although the vanilla GAN shows its potential for data generation \cite{goodfellow2014generative}, and the discriminator in this structure is proved to be effective on classification task \cite{radford2015unsupervised}. But it still suffers from blurry and possible mode dropping/collapse. Besides, there is no control in the generation process since its unsupervised manner \cite{acgan}. To this end, some studies introduce other machine learning techniques into the original GAN structure, and some results are promising. The architecture of those variants is shown in Figure~ \ref{fig:variants}.

Mirza et al. \cite{cgan} proposed CGAN (Conditional GAN), which introduces a support info vector $y$. In the generator, each input $z$ gets its corresponding $y$, and it is also available the discriminator which can help it better judge. Since this vector is a controlled parameter rather than another random sample, we gain some control of the samples generated. Chen et al. \cite{infogan}, on the other hand, is also focused on providing support info to the generator, and proposed the InfoGAN. A latent code $c$ is adding to the input of the generator, and after the images go through the discriminator, another module $Q$ is introduced to approximate the distribution of $P(c|x)$ and calculate the variational mutual information $I(c; G(z, c))$ which indicates the level of info remains after the generation process. The result generator can be controlled by maximising this regularisation term according to the latent code $c$. Odena et al. \cite{acgan} introduced a supervised task into the original GAN and proposed ACGAN (Auxiliary Classifier GAN). Every sample from the real data belongs to a predefined class, and an expected label $c\sim p_c$ along with noise $z \sim p_z$ is used as input to generate a data sample of that class. Besides the real/fake discrimination task, an auxiliary classifier is created to classify every sample, enabling the generator's ability to synthesis sample for a particular class.

\section{GANs for Spatio-temporal Data Modelling}
\label{sec:st}
\begin{figure}
    \centering
    \includegraphics[width=0.6\textwidth]{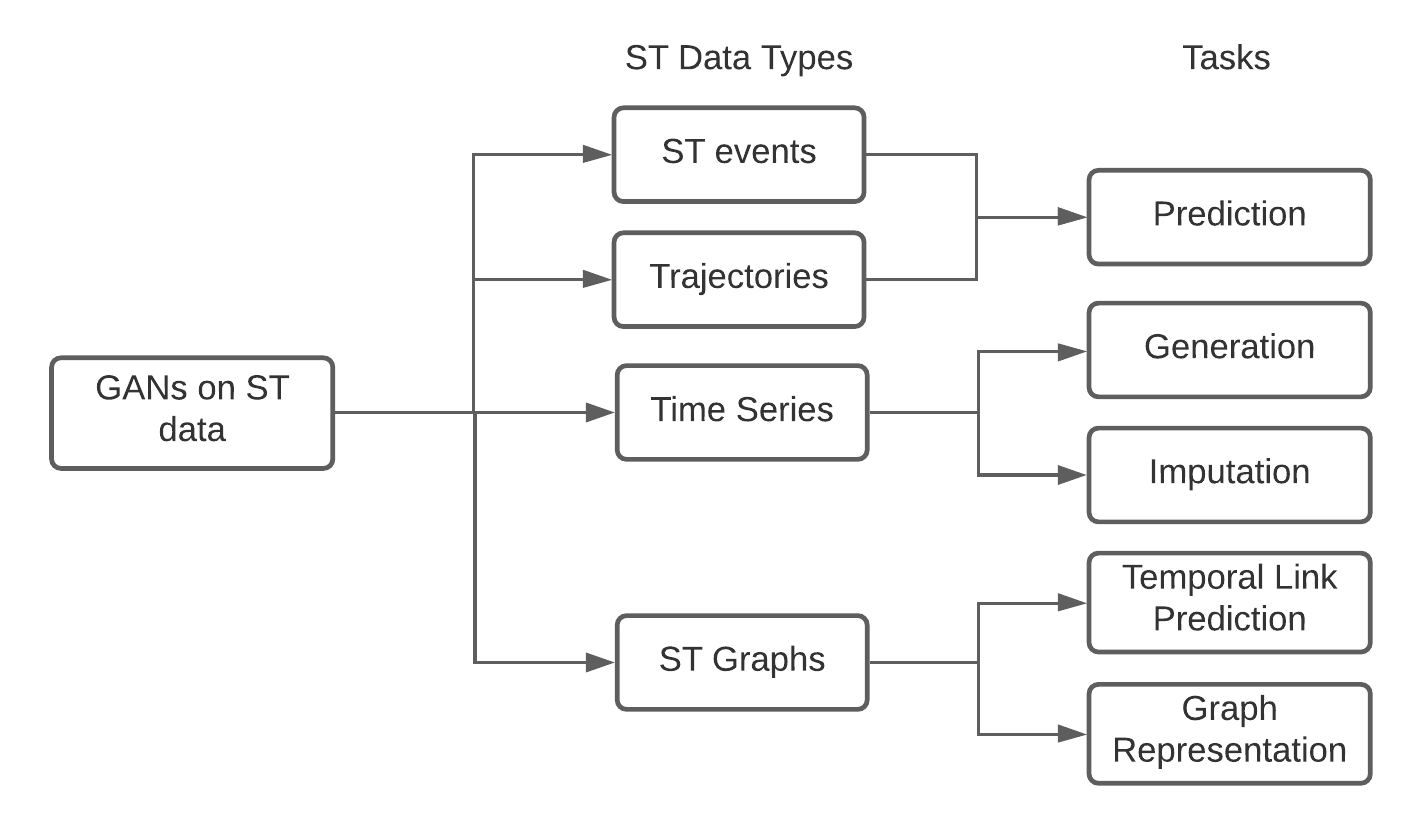}
    \caption{Taxonomy of GANs on ST data mining and modelling tasks}
    \label{fig:tax}
\end{figure}

In Figure ~\ref{fig:tax}, we categorise the existing ST data mining and modelling tasks based on four common types of ST data that have been intensively studied with GANs: ST events, time series, ST graphs and trajectories. The ST tasks are summarised based on the previous research on each ST data type (see Table~\ref{tab:category} for details). For instance, we investigate the prediction problem for the ST events and trajectories. For time-series data, we focus on the time series imputation and generation problems and for ST graphs, the temporal link prediction and graph representation applications on GANs are explored. We also summarise the widely used datasets for each ST data type in Table~\ref{tab:dataset}.

\subsection{GANs for Spatio-temporal Events}
In this subsection, we mainly introduce how GANs are applied to predict the ST events (e.g., taxi demand \cite{saxena2019d,yu2020extracting}, crime \cite{jin2019crime}, fluid flows \cite{cheng2020data}, anomaly detection \cite{li2019mad}) in the future based on the previous events.

\newgeometry{,hmargin=0.2cm,vmargin=1cm,landscape}
\begin{table}
 \centering
   \footnotesize
    \setlength\tabcolsep{3pt}
\caption{Review of Spatio-temporal Data and Modelling Tasks using GAN}
\label{tab:category}
  \centering
\begin{tabular}{@{}lllllll@{}}
\toprule
ST data type                   & Reference             & Year & Task            & Values in task                        & Model category     & Evaluation methods                                                             \\ \midrule
\multirow{15}{*}{\textit{Time series}} & C-RNN-GAN \cite{mogren2016c}    & 2016 & Generation      & Musical data                          & GAN and LSTM       & \begin{tabular}[c]{@{}l@{}}Domain metrics (e.g., polyphony, scale consistency,   repetitions, tone span)\end{tabular}\\
                              & RCGAN \cite{esteban2017real}         & 2017 & Generation      & Medical data                          & GAN and RNN        & TSTR and TRTS                                                                  \\
                              & SSL-GAN  \cite{che2017boosting}     & 2017 & Generation      & Electronic health records             & GAN, CNN and AE    & Prediction accuracy over multiple data combinations                            \\
                              & OccuGAN \cite{chen2018building}       & 2018 & Generation      & Occupancy data                        & GAN and DNN        & \begin{tabular}[c]{@{}l@{}}Domain metrics (e.g., time of first arrival, number of  occupied transitions)\end{tabular}    \\
                              & Grid-GAN \cite{zhang2018generative}   & 2018 & Generation      & Smart grid data                       & CGAN and CNN       & TSTR and TRTS                                                                  \\
                              & EEG-GAN \cite{hartmann2018eeg}     & 2018 & Generation      & EEG brain signals                     & WGAN and CNN       & IS, FID and ED                                                                 \\
                              & StockGAN \cite{zhou2018stock}        & 2018 & Generation      & Stock data                            & GAN, CNN and LSTM  & Prediction accuracy (e.g., RMSRE, DPA)                                         \\
                              & GRU-GAN \cite{luo2018multivariate}            & 2018 & Imputation      & Medical records, meteorologic data    & GAN and GRU        & Imputation accuracy                                                            \\
                              & ForGAN  \cite{koochali2019probabilistic}      & 2019 & Generation      & Synthetic series and internet traffic & CGAN and LSTM      & KL divergence                                                                  \\
                              & NAOMI   \cite{liu2019naomi}      & 2019 & Imputation      & Traffic flow, movement data           & GAN and RNN        & Imputation accuracy                                                            \\
                              & TimeGAN \cite{yoon2019time}      & 2019 & Generation      & Sines, stocks, energy and events data & GAN and AE         & Diversity, fidelity and usefulness (e.g., TSTR)                                \\
                              & E2GAN  \cite{luo20192}         & 2019 & Imputation      & Medical records, meteorologic data    & GAN and GRU        & Imputation accuracy                                                            \\
                              & SimGAN  \cite{golany2020simgans}       & 2020 & Generation      & Heart rate ECG signals                & GAN                & Prediction accuracy over multiple GAN methods                                  \\
                              & Ad-Attack  \cite{dang2020adversarial}   & 2020 & Generation      & Stock prices and electricity data     & GAN                & Domain metrics (e.g, attack sucess rate, returned  of perturbed portfolio)     \\
                               & AOS4Rec  \cite{DBLP:conf/ijcai/ZhaoSZXB20}   & 2020 & Generation      &    Sequences of recommendation  &       GAN and GRU          &   Precision, nDCG and BLEU   \\
                              \midrule
\multirow{9}{*}{\textit{Trajectory} }  & GD-GAN \cite{fernando2018gd}      & 2018 & Prediction      & Pedestrain trajectories               & GAN and LSTM       & \begin{tabular}[c]{@{}l@{}}Average displacement error (ADE) and final  displacement error (FDE)\end{tabular}               \\
                              & SocialGAN \cite{gupta2018social}      & 2018 & Prediction      & Socailly acceptable trajectories      & GAN and LSTM       & \begin{tabular}[c]{@{}l@{}}Quantitative (e.g., ADE, FDE) and qualitative  (e.g., group avoiding) metrics \end{tabular}  \\
                              & SoPhie \cite{sadeghian2019sophie}          & 2019 & Prediction      & Pedestrain trajectories               & GAN and LSTM       & ADE and FDE                                                                    \\
                              & Social Ways \cite{amirian2019social}    & 2019 & Prediction      & Pedestrain trajectories               & GAN and LSTM       & ADE and FDE                                                                    \\
                              & Social-BiGAT \cite{kosaraju2019social}     & 2019 & Prediction      & Pedestrain trajectories               & GAN and LSTM       & ADE and FDE                                                                    \\
                              
                              & APOIR  \cite{zhou2019adversarial}     & 2019 & Prediction      &   Point-of-Interests             &     GAN and GRU  &               Precision, Recall, nDCG and MAP                                    \\
                              & CoL-GAN \cite{CoLGAN}       & 2020 & Prediction      & Pedestrain trajectories               & GAN, CNN and LSTM  & Average collision times (ACT), ADE and FDE                                     \\ 
                              
                               & AdattTUL \cite{gao2020adversarial}    & 2020 & Link prediction       &       Human mobility trajectories         &            GAN, GRU and LSTM &  Prediction accuracy over multiple models                             \\
                                & MT-ASTN \cite{wang2020multi}    & 2020 & Prediction       &       Crowd flow trajectories         &  GAN, AE &        MAE and RMSE over multiple models                     \\ 
                               \midrule
\multirow{5}{*}{\textit{ST events} }   & D-GAN \cite{ren2017d}        & 2017 & Prediction      & Taxi and bike data                    & GAN and VAE        & Prediction accuracy over multiple models                                       \\
                              & Taxi-CGAN  \cite{yu2020extracting}    & 2020 & Prediction      & Taxi hotspots data                    & CGAN and LSTM      & \begin{tabular}[c]{@{}l@{}}False identification test (FIT) and the section
                              consistency test (SCT)\end{tabular}        \\
                              & Crime-GAN \cite{jin2019crime}     & 2017 & Prediction      & Crime data                            & DCGAN, CNN and RNN & \begin{tabular}[c]{@{}l@{}}Prediction accuracy (e.g., JS divergence) over multiple models\end{tabular}                   \\
                              & MAD-GAN \cite{li2019mad}     & 2019 & Prediction      & Cyber-attacks data                    & GAN and LSTM       & DR-score                                                                       \\ 
                              \midrule
\multirow{9}{*}{\textit{Graphs}  }     & GraphGAN \cite{wang2018graphgan}      & 2018 & Representation  & Social networks                       & GAN and DNN        & Prediction accuracy over multiple models                                       \\
                              & NetGAN \cite{bojchevski2018netgan}      & 2018 & Representation  & Citation and blogs networks           & WGAN and LSTM      & Prediction accuracy over multiple models                                       \\
                              & ANE  \cite{dai2018adversarial}          & 2018 & Representation  & Citation and blogs networks           & GAN and DNN        & Prediction accuracy over multiple models                                       \\
                              & NetRA \cite{yu2018learning}   & 2018 & Representation  & Social and biological networks        & GAN, LSTM and AE   & Prediction accuracy over multiple models                                       \\
                              & GCN-GAN \cite{lei2019gcn}     & 2019 & Link Prediction & Mobility networks                     & GAN, GCN and LSTM  & MSE, edge-wise KL divergence, mismatch rate                                    \\
                              & GANE \cite{hong2019gane}         & 2019 & Representation  & Coauthor networks                     & WGAN and DNN       & Prediction accuracy over multiple models                                       \\
                              & NetworkGAN  \cite{yang2019advanced}  & 2019 & Link Prediction & Social networks                       & GAN, GCN and LSTM  & RMSE, AUC, KL divergence                                                       \\
                              & ProGAN  \cite{gao2019progan}       & 2019 & Representation  & Social and citation networks          & GAN and DNN        & Prediction accuracy over multiple models                                       \\
                              & MEGAN \cite{sun2019megan}        & 2019 & Representation  & Social multi-view networks            & GAN and MLP        & Prediction accuracy over multiple models    \\  
                                 & GRL \cite{wang2020grl}    & 2020 & Link Prediction       &      Relation triples and freebase entity pairs      &  GAN, LSTM and RL &        MAE, MAP and AUC over multiple models                                                   \\ \bottomrule 
\end{tabular}
\end{table}
\restoregeometry

\begin{figure}
    \centering    \includegraphics[width=0.95\textwidth]{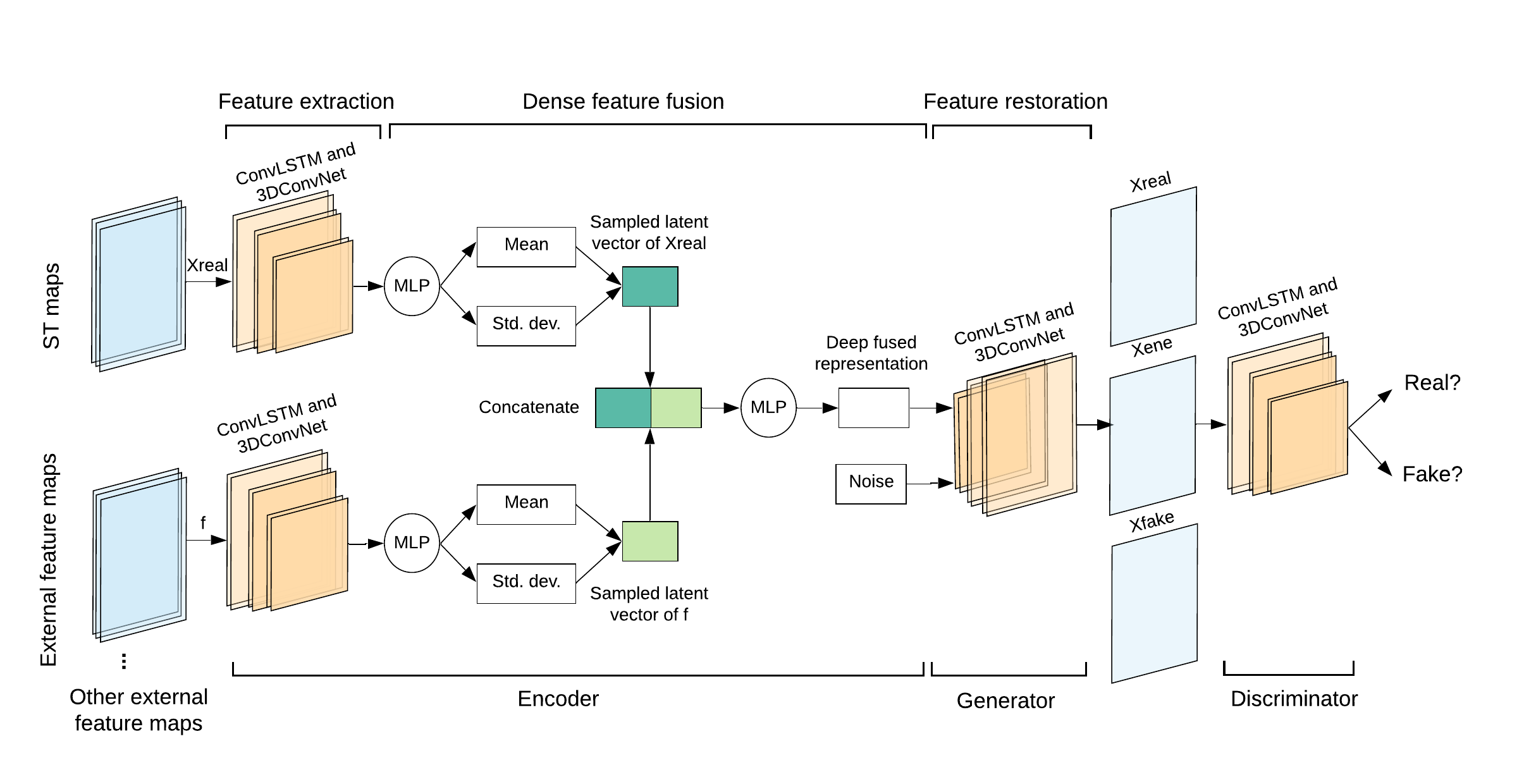}
    \caption{D-GAN architecture proposed by Seaena et al. \cite{saxena2019d}}
    \label{fig:dgan}
\end{figure}

For the first time, Saxena et al. \cite{saxena2019d} proposed a generative adversarial network \textit{D-GAN} for accurate spatio-temporal events prediction. In the model, GAN and VAE are combined to jointly learn generation and variational inference of ST data in an unsupervised manner. They also designed a general fusion module to fuse heterogeneous multiple data sources. Figure~\ref{fig:dgan} shows the architecture for D-GAN, consisting of four components: \textit{Encoder}, \textit{Generator/Decoder}, \textit{Discriminator}, and \textit{External feature fusion}. $G$ network is trained using the adversarial process. The decoder (i.e., generator) learns to approximate the distribution of real data, while the $D$ network discriminates between samples generated by $D$ and samples from real distributions. During the training process, D-GAN adopts a reconstruction loss and adversarial loss \cite{saxena2019d}. In addition, \textit{ConvLSTM} \cite{xingjian2015convolutional} and \textit{3D
-ConvNet} structures were exploited to model long-term patterns and spatial dependencies in ST data.

Recently, Yu et al. \cite{yu2020extracting} applied a conditional generative adversarial network with long short-term structure (LSTM-CGAN) for taxi hotspot prediction, which captures the spatial and temporal variations of hotspots simultaneously. Furthermore, Jin et al. \cite{jin2019crime} developed a context-based generative model \textit{Crime-GAN} to learn the spatio-temporal dynamics of the crime situation. They aggregated Seq2Seq, VAE network and adversarial loss in the framework to study ST data representation better. Furthermore, the deep convolutional generative adversarial network (DCGAN) has been developed for spatio-temporal fluid flow prediction in a tsunami case in Japan \cite{cheng2020data}.

GANs have also been used for anomaly detection for ST events. Li et al. \cite{li2019mad} proposed MAD-GAN, an unsupervised anomaly detection method for multivariate time series based on GAN. They trained a GAN generator and discriminator with LSTM. Then, the GAN-trained generator and discriminator are employed to detect anomalies in the testing data with a combined Discrimination and Reconstruction Anomaly Score (DR-Score).

\subsection{GANs for Trajectory Prediction}
Trajectory prediction refers to the problem of estimating the future trajectories of various agents based on the previous observations \cite{ma2019trafficpredict}. Gupta et al. \cite{gupta2018social} proposed {SocialGAN} to jointly predict trajectories avoiding collisions for all people. They introduced a variety loss encouraging the generative network of the GAN to spread its distribution and cover the space of possible paths while being consistent with the observed inputs. A new pooling mechanism was proposed to learn a 'global' pooling vector that encodes the subtle cues for all people involved in a scene. In {GD-GAN}~\cite{fernando2018gd}, Fernando et al. designed a GAN based pipeline to jointly learn features for both pedestrian trajectory prediction and social group detection. As the basic GAN structure used in SocialGAN is susceptible to mode collapsing and dropping issues, Amirian et al.~\cite{amirian2019social} extended the SocialGAN by incorporating the Info-GAN~\cite{infogan} structure in their \textit{Social Ways} trajectory prediction network.

\textit{SoPhie}, proposed by Sadeghian et al.~\cite{sadeghian2019sophie}, is another GAN based trajectory prediction approach that can take both the information from the scene context and social interactions of the agents into consideration.
Two separate attention modules are also used to better capture the scene context and the social interactions.
More recently, based on BicycleGAN~\cite{zhu2017toward} framework, Social-BiGAT~\cite{kosaraju2019social} develops the bijection function between the output trajectories and the latent space input to the trajectory generator.
It also uses a Graph Attention Network in combination with a VGG network~\cite{simonyan2015very} to encode social influence from other pedestrians and semantic scene influence of the environment. For trajectories with fewer potential collisions, CoL-GAN~\cite{CoLGAN}, proposed by Liu et al., exploits a CNN-based network as the trajectory discriminator.
Different from other GAN based trajectory prediction methods such as SocialGAN~\cite{gupta2018social} and SoPhie~\cite{sadeghian2019sophie}, the proposed discriminator can classify whether each segment of a trajectory is real or fake.

Recently, Gao et al. \cite{gao2020adversarial} studied the trajectory user linking problem to identify user identities from mobility
patterns. They combined autoencoder with GANs for jointly human mobility learning, which provides regularized latent space for mobility classification. APOIR \cite{zhou2019adversarial} was developed to learn the distribution of underlying user preferences in the Point-of-interest (POI) recommendation. It consists of two components: the recommender and the discriminator. The recommender approaches users' true preference, and the discriminator distinguishes the generated POIs from the truly visited ones.


\subsection{GANs for Time Series Modelling}

Specifically, time series is assumed to be a special kind of sequential data where the order matters. It is a sequence series obtained at consecutive equally spaced points at the time dimension, and not the only case of sequential data. However, processing the sequential data (e.g., musical data) may share similar GAN approaches to the time series data. Therefore, we will include several studies for modelling sequential data (e.g., musical data in \cite{mogren2016c}). Although natural language data can also be considered as sequential data, we will not include the NLP research with GANs (e.g., text generation \cite{lin2017adversarial}, text to image \cite{zhang2017stackgan}) since the natural language is not one of the ST data types and usually falls into the field of NLP. In this subsection, we will demonstrate two ST tasks for time series data: generation and imputation.

\subsubsection{Generation}
Data generation refers to creating data from the sampled data source. One of the main purposes of time series generation with GAN is to protect the privacy of sensitive data such as medical data \cite{esteban2017real}, electroencephalographic  (EEG) data \cite{hartmann2018eeg},  heart signal electrocardiogram (ECG) data \cite{golany2020simgans}, occupancy data \cite{chen2018building}, electronic health records (EHR) \cite{che2017boosting}, etc.


Recently, GANs have been used to generate sequential data. Mogren et al. \cite{mogren2016c} proposed C-RNN-GAN (continuous RNN-GAN) to generate continuous-valued sequential data. They built the GAN with an LSTM generator and discriminator. The discriminator consists of a bidirectional layout which allows it to take context in both directions into account for its decisions. They trained the model on sequences of classical music and evaluated with metrics such as polyphony, scale consistency, repetitions and tone span. 

Then, Esteban et al. \cite{esteban2017real} proposed a regular GAN where recurrent neural networks have substituted both the generator and the discriminator. They presented the Recurrent GAN (RGAN) and Recurrent Conditional GAN (RCGAN) to generate sequences of real-valued medical data or data subject to some conditional inputs. For evaluation, they proposed to use the capability of the generated synthetic data to train supervised models, i.e., TSTR (train on synthetic, test on real). They addressed that TSTR is more effective than TRTS (train on real, test on synthetic) because TRTS performance may not degrade when GAN suffers mode collapse.



GANs have been used for the generation of biological-physiological signals such as EEG and ECG.  Hartmann et al. \cite{hartmann2018eeg} proposed EEG-GAN to generate electroencephalographic (EEG) brain signals. With the modification of the improved WGAN training, they trained a GAN to produce artificial signals in a stable fashion which strongly resembles single-channel real EEG signals in the time and frequency domain. For evaluation metrics, they showed that the combination of Frechet inception distance (FID) and sliced Wasserstein distance (SWD), Euclidean distance (ED) can give a good idea about its overall properties. Golany et al. \cite{golany2020simgans} proposed the simulator-based GANs for ECG synthesis to improve a supervised classification. They incorporated ECG simulator equations into the generation networks, and then the generated ECG signals are used to train a deep network.

Chen et al. \cite{chen2018building} proposed GAN framework for building occupancy modelling. They first learned the discriminator and generator in the vanilla GAN with the training occupancy data. Then, the learned generator is the required occupancy model, which can be used to generate occupancy data with random inputs. To evaluate, they defined five variables (i.e., mean occupancy, time of the first arrival, time of the last departure, cumulative occupied duration and the number of occupied/unoccupied transitions) with two criteria (i.e., normalised root mean squared error and total variation distance).

Che et al. \cite{che2017boosting} used a modified GAN called \textit{ehrGAN} to generate plausible labelled EHR data. The generator is a modified encoder-decoder CNN network, and the generated EHR data mimics the real patient records which augments the training dataset in a semi-supervised learning manner. In this work, they used the generative networks with the CNN prediction model to improve the performance of risk prediction.

Koochali et al. \cite{koochali2019probabilistic} proposed \textit{ForGAN} to predict the next-step time series value $X_{t+1}$ by learning the full conditional probability distribution. They applied a conditional GAN and the condition windows are the previous $t$ values  ($X_0, X_1,...,X_t$). With the input of the noise vector, the generator predicts the values at the $t+1$ step and then the discriminator compared this value to the true value at the $t+1$ step with the same condition windows. LSTM network is used in both generator and discriminator. Zhou et al. \cite{zhou2018stock} predicted the stock price at next time step $y_{t+1}$ based on the features in previous t time step $X_1,X_2,...,X_t$ and previous stock price $y_1, y_2,...,y_t$ using generative adversarial nets. 

Instead of generating a sequence of single values, Dang et al. \cite{dang2020adversarial} developed an approach for the generation of adversarial attacks where the output is a sequence of probability distributions. The proposed approaches are demonstrated on two challenging tasks: the prediction of electricity consumption and stock market trading. Besides, AOSeRec \cite{DBLP:conf/ijcai/ZhaoSZXB20} were proposed to generate a sequence of items consistent with user preferences rather than the next-item prediction. The model integrated the sequence-level oracle and adversarial learning into the seq2seq auto-regressive learning.


Generally, an excellent time-series generative model should preserve temporal dynamics, and the generated sequences should follow the original patterns between variables across time. Therefore, Yoon et al. \cite{yoon2019time} proposed a framework \textit{TimeGAN} for producing realistic multivariate time-series, combining the flexibility of the unsupervised GAN approach with the control afforded by supervised learning. In addition to the traditional unsupervised adversarial loss on both real and fake data, they presented a stepwise supervised loss with the original data as supervision, which helps learn from the transition dynamics in real sequences. 

\begin{figure}
    \centering
    \includegraphics[width=0.9\textwidth]{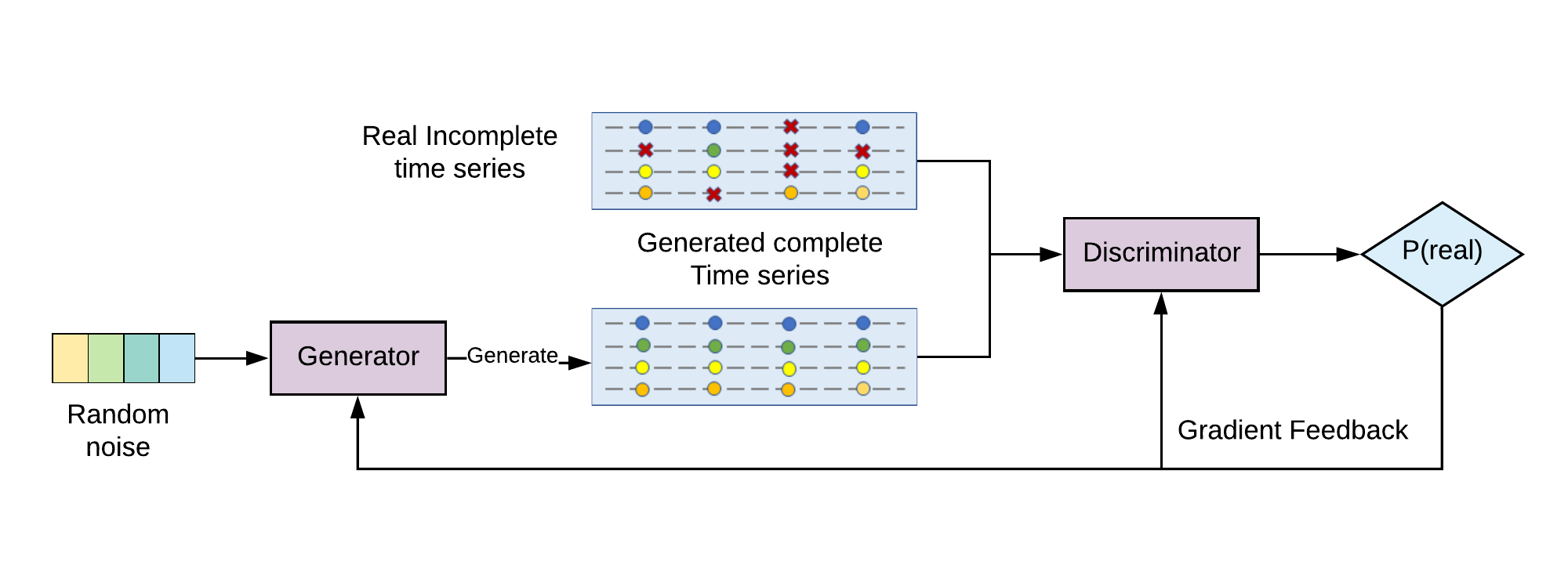}
    \caption{An overview of the time series imputation framework proposed by Luo et al. \cite{luo2018multivariate}}
    \label{fig:GANE1}
\end{figure}

\subsubsection{Imputation} In real-world applications, time series are usually incomplete due to various reasons, and the time intervals of observations are usually not fixed \cite{luo2018multivariate}. The missing values in time series make it hard for effective analysis \cite{garcia2015missing}. One popular way to handle the missing values of time series is to impute the missing values to get the complete dataset. Generally, there are three different ways for time series imputation: case deletion methods \cite{kaiser2014dealing}, statistical imputation methods \cite{graham2009missing}, and machine learning based imputation methods \cite{batista2003analysis}. However, all the existing approaches hardly consider the temporal relations between two observations. In recent years, researchers have started to take advantages of GANs to learn latent representations between observations for time series imputation \cite{luo2018multivariate,luo20192,liu2019naomi}.

Luo et al. \cite{luo2018multivariate} applied the adversarial model to generate and impute the original incomplete time series. To learn the latent relationships between observations with non-fixed time lags, a novel RNN cell called GRUI was proposed, which considers the non-fixed time lags and fades the influence of the past observations determined by the time lags. They proposed a two-stage model  (see Figure~ \ref{fig:GANE1}) for time series imputation: In the first stage, they adopted the GRUI in the discriminator and generator in GAN to learn the distribution and temporal information of the dataset. In the second stage, for each sample, they tried to optimise the 'noise' input vector and find the best-matched input vector of the generator. The noise was trained with a two-part loss function: masked reconstruction loss and discriminative loss. Masked reconstruction loss is the masked squared errors of the non-missing part between the original and generated sample. It means that the generated time series should be close enough to the original incomplete time series. The discriminative loss forces the generated sample as real as possible. However, this two-stage model needs a considerable time to find the best-matched input vector, which is not always the best, especially when the initial value of the 'noise' is not set properly. 

\begin{figure}
    \centering
    \includegraphics[width=0.9\textwidth]{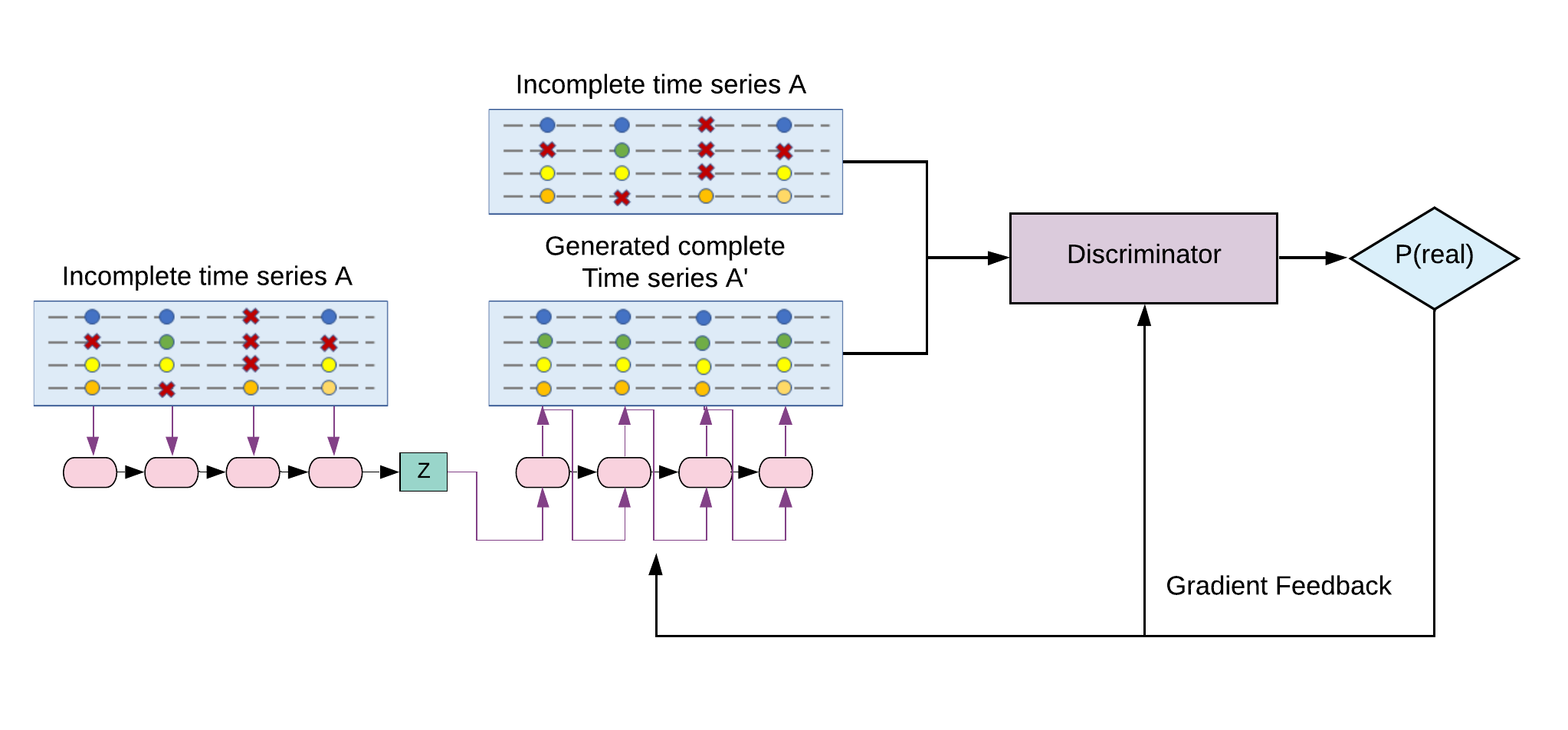}
    \caption{An overview of E$^2$GAN framework proposed by Luo et al. \cite{luo20192}}
    \label{fig:GANE2}
\end{figure}

Then, Luo et al. \cite{luo20192} proposed an end-to-end GAN-based imputation model E$^2$GAN which not only simplifies the process of time series imputation but also generates more reasonable values for the filling of missing values. E$^2$GAN takes a compressing and reconstructing strategy to avoid the 'noise' optimisation stage in \cite{luo2018multivariate}. As seen in Figure ~\ref{fig:GANE2}, in the generator (a denoising auto-encoder), they added a random vector to the original sample and map it into a low-dimensional vector. Then they reconstructed it from the low-dimensional vector. The generator seeks to find a network structure that can best compress and reconstruct the multivariate time series and fool the discriminator. Then they used the reconstructed sample to impute the missing values.

Non-Autoregressive Multiresolution Imputation (NAOMI) \cite{liu2019naomi} is a new model for the imputation of spatio-temporal sequences like traffic flow data and movement trajectories when arbitrary missing observations are given. NAOMI imputes missing values for spatio-temporal sequences recursively from coarse to fine-grained resolutions with a non-autoregressive decoding procedure. It further employs a generative adversarial learning process to reduce variance for improving the performance.

\subsection{GANs for Spatio-temporal Graph Modelling}

In this subsection, we introduce the application of GAN on the graph data analysis which mainly focus on two areas: temporal link prediction and graph representation.

\subsubsection{Temporal Link Prediction} Temporal link prediction refers to the dynamics prediction problem in network systems (e.g., mobility and traffic prediction) where system behaviours are described by the abstract graphs \cite{lei2019gcn}. Given the snapshots of a graph in previous timestamps, the temporal link prediction task aims to construct the graph topology at the next timestamp. Lei et al. \cite{lei2019gcn} proposed GCN-GAN to predict links in weighted dynamic networks. They combined graph convolutional network (GCN), long short-term memory (LSTM) as well as generative adversarial network (GAN). The generator consists of a GCN hidden layer, LSTM hidden layer and a fully connected layer. Discriminator contains a fully connected feed-forward network. For evaluation, they used edge-wise KL divergence and mismatch rate besides mean square error (MSE). Then, Yang et al. \cite{yang2019advanced} designed an attentive GCN model for temporal link prediction in graphs using GAN. Compared to \cite{lei2019gcn}, attentive GCN allows for assigning different importance to the vertices to learn the spatial features of the dynamic network. Then, temporal matrix factorisation (TMF) LSTM was employed to capture dynamic networks' temporal dependencies and evolutionary patterns. GAN framework was then proposed to improve the performance of temporal link prediction. 

Recently, Want et al. \cite{wang2020grl} have designed a GAN-based reinforcement learning model (GRL) for knowledge graph completion, which employs both WGAN and LSTM to record trajectories and generate sub-graph sequences. In addition, the deep deterministic policy gradient approach (DDPG) is adopted to optimise both reward and adversarial loss and generates better policies, which leads to more stable training compared with the traditional optimization method.   

\subsubsection{Graph Representation}
Wang et al. \cite{wang2018graphgan} proposed GraphGAN unifying two types of graph representation methods: discriminative methods and generative methods via adversarial training. They found that the traditional softmax function and its variants are not suitable for the generator for two reasons: (1) softmax treats all vertices equally in the graph for a given vertex and does not consider the graph structure and proximity information; (2) the calculation of softmax involves all vertices in the
graph which is time-consuming and computationally inefficient. Therefore, they introduced \textit{graph softmax} as the implementation of the generator and proved that it satisfies the desirable properties of normalisation, computational efficiency and graph structure awareness.


\begin{table}
\small
\setlength\tabcolsep{2pt}
\caption{Summary of Datasets}
\label{tab:dataset}
\begin{tabular}{@{}llll@{}}
\toprule
ST data type                  & Dataset                                                                 & Data source                                       & Used references                                                       \\ \midrule
\multirow{10}{*}{\textit{Time series}} & Philips eICU database \cite{pollard2018eicu}           & Medical data  & \cite{esteban2017real}                                                         \\
                              & MNIST             \cite{xiao2017fashion}                                                       & Hand-written digit images                         & \cite{esteban2017real}                                                         \\
                              & Occupancy dataste \cite{liao2012agent}                 & Occupancy data in the building                    & \cite{chen2018building}                                                     \\
                              & Pecan street dataset \cite{johnson_2020}              & Energy consumption, solar generation           & \cite{zhang2018generative}                                                   \\
                              & PhysioNet dataset \cite{silva2012predicting}           & Medical data (e.g., heart rate, glucose)          & \cite{golany2020simgans,luo2018multivariate, luo20192}                    \\
                              & KDD cup 2018 dataset \cite{kdd2018}                    & Air quality data                                  & \cite{luo2018multivariate,luo20192}                                     \\
                              & A5M dataset \cite{cortez2012multi}                     & Transatlantic link data                           & \cite{koochali2019probabilistic}                                                  \\
                              & PEMS-SF traffic dataset \cite{dua2017uci}              & Freeway occupancy rate                            &   \cite{liu2019naomi}                                                    \\
                              & Appliances energy dataset \cite{candanedo2017data} & Environmental data                                & \cite{yoon2019time}                                                          \\ 
                              & UCI electricity dataset \cite{trindade2016uci}         & Historical price data                             & \cite{dang2020adversarial}                \\ 
                              &  Yoochoose \cite{ben2015recsys}        &   Clicking events from users                         &    \cite{DBLP:conf/ijcai/ZhaoSZXB20}                                                            \\ 
                               & MovieLens \cite{harper2015movielens}        &   Movie ratings data                          &      \cite{DBLP:conf/ijcai/ZhaoSZXB20}                                                                \\ 
                               \midrule
\multirow{8}{*}{\textit{Trajectory}}   & ETH     \cite{pellegrini2010improving}                                                                & Videos                          &       \cite{fernando2018gd,CoLGAN,gupta2018social,amirian2019social,sadeghian2019sophie,kosaraju2019social}                                                                \\
                              & UCY     \cite{lerner2007crowds}                                                                & Videos                           &        \cite{fernando2018gd,CoLGAN,gupta2018social,amirian2019social,sadeghian2019sophie,kosaraju2019social}                                                               \\
                              & Stanford drone dataset   \cite{robicquet2016learning}                                         & Videos                    &    \cite{sadeghian2019sophie}                                                                   \\
                              & Vittorio emanuele II             \cite{bandini2014towards}           & Videos                          &    \cite{fernando2018gd}                                                                                \\
                              & Foursquare \cite{yang2014modeling}         &         Location-based social networks               &    \cite{gao2020adversarial, zhou2019adversarial,liu2019geo,manotumruksa2020sequential,li2020adversarial}                                                                              \\
                              
                               & Gowalla \cite{cho2011friendship}     & Location-based social networks                         &    \cite{gao2020adversarial, zhou2019adversarial,liu2019geo}                                                                              \\
                               & Brightkite    \cite{cho2011friendship}     &      Location-based social networks                   &    \cite{gao2020adversarial,manotumruksa2020sequential}                                                                              \\
                               & Yelp  \cite{yelpdataset}      &  Location-based social networks                       &    \cite{zhou2019adversarial,manotumruksa2020sequential,li2020adversarial}                                                                              \\
                              \midrule
\multirow{3}{*}{\textit{ST events}}    & Yellow taxi dataset \cite{tlctrip}                     & Taxi demand data                                  & \cite{ren2017d}                                                       \\
                              & CitiBike trip dataset \cite{citibike}                  & Bike demand data                                  & \cite{ren2017d}                                                       \\
                              & SWaT dataset \cite{goh2016dataset}                     & Attacked data in water system                     & \cite{li2019mad}                                            \\ \midrule
\multirow{8}{*}{\textit{Graphs}}       & ArXiv-AstroPh \cite{snap}                              & Scientific collaborations data                    & \cite{wang2018graphgan}                                                    \\
                              & Wikipedia \cite{grover2016node2vec}            & Network of words                                  &
                              \cite{dai2018adversarial, wang2018graphgan, yu2018learning}               \\
                              & CORA \cite{cora}                               & Citation networks of publications                 &\cite{bojchevski2018netgan, dai2018adversarial, gao2019progan}                      \\
                              & CiteSeer \cite{sen2008collective}              & Citation networks of publications                 & \cite{bojchevski2018netgan,  dai2018adversarial,  gao2019progan}                       \\
                              & DBLP \cite{pan2016tri}                         & Collaboration graph of authors                    & \cite{bojchevski2018netgan,  dai2018adversarial,   yang2019advanced, yu2018learning}  \\
                              & Blogcatalog \cite{tang2009relational}                  & Social network for bloggers                       & \cite{gao2019progan, wang2018graphgan,  yu2018learning}                  \\
                              & UCI message dataset \cite{ucimessage}                  & Message communication networks                    & 
                              \cite{yang2019advanced, yu2018learning}                               \\
                              & Flickr \cite{nr}                               & Social networks                                   &  \cite{gao2019progan,sun2019megan}                                      \\ \bottomrule
\end{tabular}
\end{table}

Aiming at better capturing the essential properties and preserving the patterns of real graphs, Bojchevski et al. introduced NetGAN \cite{bojchevski2018netgan} to learn a distribution of network via the random walks. The merits of using random walks are their invariance under node reordering and efficiency in exploring the sparse graphs by merely traversing the nonzero entries. The results confirmed that the combination of longer random walks and LSTM is advantageous for the model to learn the topology and general patterns in the data. 

Adversarial Network Embedding (ANE) \cite{dai2018adversarial} also considers the random walk mechanism to learn network representation with the adversarial learning principle. It consisted of two components: (1) the structure-preserving component is developed to extract network structural properties via either Inductive DeepWalk or Denoising Autoencoder; (2) the adversarial learning component contributes to learning network representations by matching the posterior distribution of the latent representations to given priors. However, using DeepWalk for learning graph embedding could lead to an overfitting issue due to sparsity is common in networks or increasing computational burden when more sampled walks are considered \cite{yu2018learning}. Therefore, NetRA \cite{yu2018learning} was proposed to further minimise network locality-preserving loss and global reconstruction error with a discrete LSTM Autoencoder and continuous space generator, such that the mapping from input sequences into vertex representations could be improved.

Most recently, GAN embedding (GANE) \cite{hong2019gane} tries to gain the underlying graph distribution based on the probability distribution of edge existence which is similar to GraphGAN. The difference is that this model applies Wasserstein-1 distance as the overall objective function and intends to achieve link prediction and network embedding extraction simultaneously. As a novel network embedding method, the proximity generative adversarial network (ProGAN) \cite{gao2019progan} is proposed to capture the underlying proximity between different nodes by approximating the generated distribution of nodes in a triplet format to the underlying proximity in the model of GAN. Specifically, a triplet can encode the relationship among three nodes, including similarity and dissimilarity. After the training of the generator and discriminator, the underlying proximities discovered are then used to build network embedding with an encoder.

The works mentioned above primarily focus on the single-view network in learning network embedding. However, numerous real-world data are represented by multi-view networks whose nodes have different types of relations. Sun et al. \cite{sun2019megan} introduced a new framework for multi-view network embedding called MEGAN, which can preserve the information from individual network views, while considering nodes connectivity within one relation and complex correlations among different views. During the training of MEGAN, a pair of nodes are chosen from the generator based on the fake connectivity pattern across views produced by multi-layer perceptron (MLP), and the discriminator is then executed to differentiate the real pair of nodes from the generated one. 

\section{Discussion}
\label{sec:dis}


\subsection{Challenges and Future Directions}

Alongside numerous advantages of GANs, there are still challenges needed to be solved for employing GANs for ST applications.
The traditional architectures and loss functions of GANs may not be suitable due to the unique properties of ST data. Besides, evaluating ST data is more difficult compared to images where researchers could rely on visual inspections. Therefore, we will mainly focus on: (1) \textit{how to modify architectures/loss functions of GANs to better capture the spatial and temporal relations for ST data and achieve stable training?} (2) \textit{how to evaluate the performance of GANs especially when visually inspecting the generated ST samples is not applicable?} We will then address these two problems and indicate the future directions of investigating this area.

\subsubsection{Architectures and loss functions of GANs} 

In the computer vision area, fully connected layers were initially used as building blocks in vanilla GAN, but later on were replaced by convolutional layers in DCGAN \cite{radford2015unsupervised}. Compared with images with only spatial relations, modelling ST data is more complex due to the constraints from both spatial and temporal dimensions. Therefore, adapting architectures and loss functions of GANs for specific ST applications have become the mainstream recently.

Generally, original or adapted RNN  \cite{mogren2016c,esteban2017real,luo2018multivariate} , LSTM \cite{li2019mad,koochali2019probabilistic,lei2019gcn,bojchevski2018netgan,yu2018learning}, VAE \cite{saxena2019d,che2017boosting,luo20192,yu2018learning}, CNN \cite{che2017boosting}, GNN \cite{lei2019gcn} are usually used as the base model (i.e., the discriminator and generator) in the vanilla GAN , WGAN \cite{hartmann2018eeg} or CGAN \cite{koochali2019probabilistic}, which captures the spatio-temporal relations for ST data. What's more, some new loss functions have been proposed to dealing with specific ST tasks, such as the stepwised supervised loss in TimeGAN \cite{yoon2019time}, masked reconstruction loss in GRU-GAN \cite{luo2018multivariate}, the variety loss in SocialGAN \cite{amirian2019social}. 

The architecture of the generator and discriminator is of significant importance since it strongly influences the performance and stability of the GANs on ST data. Though GAN models have achieved remarkable success in ST applications \cite{saxena2019d,yu2020extracting,jin2019crime,cheng2020data,li2019mad}, the unstable training process still remains unresolved and hinders further development for GAN on ST tasks, especially considering the heterogeneity and auto-correlation of ST data. For instance, Saxena et al. \cite{saxena2019d} concatenated the latent code and data space in the discriminator for faster convergence, better learning and higher training stability. Although many previous studies discussed how to enable the stable training process \cite{che2016mode,salimans2016improved,gulrajani2017improved}, the problems of instability of GANs still need further research, especially on the ST data modelling. With further developments of GANs, new architectures and loss functions can be designed based on the characteristics of ST tasks.

\subsubsection{Evaluation Metrics}
Though GANs have gained huge success in various fields, evaluating the performance of GANs is still an open question. As illustrated in \cite{borji2019pros} and \cite{hong2019generative}, both quantitatively measures (e.g., \textit{Log-likelihood with Parzen Window Estimation} \cite{salimans2016improved}, \textit{Fréchet Inception Distance} \cite{heusel2017gans}, \textit{Maximum Mean Discrepancy} \cite{gretton2012kernel}, \textit{Root Mean Square Error} \cite{r2mse}, \textit{Histogram} \cite{hisgan}, \textit{Stepwise Method} \cite{stepwise}) and qualitative measures (e.g., \textit{Preference Judgement} \cite{wang2018stacked}, Analysing \textit{Internals of Models} \cite{radford2015unsupervised}) have strengths and limitations. The nebulous notion of quality can be best assessed by a human judge, which is neither practical nor appropriate for different types of ST data. 

In most cases, it is not easy or even possible to visually evaluate the generated ST data. For instance, the \textit{Intense Care Unit} (ICU) time series \cite{esteban2017real} or heart rate \textit{Electrocardiogram} (ECG) \cite{golany2020simgans} signals could look completely random to a non-medical expert. Usually, the evaluation of generated ST samples requires domain knowledge. For example, Mogren et al. \cite{mogren2016c} evaluated the generated music sequences using metrics in the field of music such as polyphony, repetitions, tone span and scale consistency. For future ST applications with GANs, some novel metrics based on domain knowledge could be considered to evaluate the generated ST data. 

Especially, some researchers have proposed the general approach to evaluate the generated ST-data. Esteban et al. \cite{esteban2017real} developed a general method called \textit{Train on Synthetic, Test on Real} (TSTR) to evaluate the generated samples of GANs when a supervised task defined on the training data. They used a dataset generated by GANs to train a classification model, then tested on a held-out set of true samples. This evaluation metric is ideal when employing GANs to share synthetic de-identified data because it demonstrates the ability of the generated synthetic data to be used for real applications. In the future, more practical metrics should be developed to evaluate the performance of generated ST samples.

\subsection{Conclusions}
In this survey, we conducted a comprehensive overview of \textit{Generative Adversarial Networks} (GANs) for spatio-temporal (ST) data in recent years. Firstly, we discussed the properties of ST data and traditional ways for ST data modelling. Then, we have provided a thorough review and comparison of the popular variants of GAN, and its applications on ST data analysis, such as time series imputation, trajectory prediction, graph representation and link prediction. Besides, we summarised the challenges and future directions for employing GANs for ST applications. 

Finally, though there are many promising results in the literature, we would like to point out, the adoption of GANs for ST data is still in its infancy. This survey can be used as the stepping stone for future research in this direction, which provides a detailed explanation of different ST applications with GANs.  We wish this paper could help readers identify the set of problems and choose the relevant GAN techniques when given a new ST dataset.

\section{Acknowledgments}
This research was supported by the Australian Government through the Australian Research Council's Linkage Projects funding scheme (LP150100246) and Discovery Project (DP190101485). We also acknowledge the support of RMIT Research Stipend Scholarship and CSIRO Data61 Scholarship.

\bibliographystyle{ACM-Reference-Format}
\bibliography{nan.bib}

\end{document}